# [1]Date-Field Retrieval in Scene Image and Video Frames using Text Enhancement and Shape Coding

[a]Partha Pratim Roy*, [b]Ayan Kumar Bhunia, [c]Umapada Pal

[a]Dept. of CSE, Indian Institute of Technology Roorkee, India
[b]Dept. of ECE, Institute of Engineering & Management, Kolkata, India
[c]CVPR Unit, Indian Statistical Institute, Kolkata, India
[a]email: proy.fcs@iitr.ac.in, TEL: +91-1332-284816

## Abstract

Text recognition in scene image and video frames is difficult because of low resolution, blur, background noise, etc. Since traditional OCRs do not perform well in such images, information retrieval using keywords could be an alternative way to index/retrieve such text information. Date is a useful piece of information which has various applications including date-wise videos/scene searching, indexing or retrieval. This paper presents a date spotting based information retrieval system for natural scene image and video frames where text appears with complex backgrounds. We propose a line based date spotting approach using Hidden Markov Model (HMM) which is used to detect the date information in a given text. Different date models are searched from a line without segmenting characters or words. Given a text line image in RGB, we apply an efficient gray image conversion to enhance the text information. Wavelet decomposition and gradient sub-bands are used to enhance text information in gray scale. Next, Pyramid Histogram of Oriented Gradient (PHOG) feature has been extracted from gray image and binary images for date-spotting framework. Binary and gray image features are combined by MLP based Tandem approach. Finally, to boost the performance further, a shape coding based scheme is used to combine the similar shape characters in same class during word spotting. For our experiment, three different date models have been constructed to search similar date information having numeric dates that contains numeral values and punctuations and semi-numeric that contains dates with numerals along with months in scene/video text. We have tested our system on 1648 text lines and the results show the effectiveness of our proposed date spotting approach.

---

## 1. Introduction

Nowadays, there is a huge collection of digital information in form of video or scene images available due to easy availability of mobile devices and internet communication. This creates a very large database of videos, images and documents every day, which requires proper indexing for effective management of the database [27, 42, 43]. Retrieval of relevant videos/images from this collection is time consuming and needs a lot of manual effort. Thus, text plays an important role in automatic video indexing and retrieval. Querying with text information is required to index or search image/video in datasets [44, 45]. The information could be like keywords, names, locations, date, time, etc. Date is a useful piece of information and it could be used as a key in various applications such as date-based image/video frame retrieval, searching and indexing of digital information repositories consisting of a collection of documents, images or videos. Also, date can be used to sort the documents in an archive. Often date and timestamp are added to photos in a gallery. A query with date will undoubtedly improve the searching performance from this collection. Automatic spotting or searching of date information involves challenges due to different date patterns, format lengths, punctuations and font styles etc. A date field can be numeral, which consists of only numerals and punctuations, (e.g. 22/03/2006, 10-11-15, 6.5.1994 etc.) where the date pattern follows a generalized format such as dd/mm/yyyy, mm-dd-yyyy, d/m/yyyy etc. and separated by punctuations (e.g. slash '/', hyphen '-' and period '.') or semi-numeric that consists of month-word (e.g January, Feb, May, Sept etc.), numerals and ordinal number suffix (e.g. st, nd, rd, th). Few examples of such semi-numeric format are 22nd March, 2006, March 2, 2010, Oct $2^{nd}$ 1980 etc.

Date based pattern searching has been addressed by Mandal et al. [18], but there exist no work on such indexing techniques in image/video frames. The major difficulties in recognizing the in-video texts are the low-resolution, noisy background image, immense variations in color, etc. There exist some commercial Optical Character Recognizers (OCR) available for recognizing texts in documents but these OCRs do not work properly because of the poor quality of image/videos. Recently there exists a lot of research work on text recognition in natural scene images/video frames [1, 15, 16] but the recognition accuracy is not satisfactory. Hence, we borrow the "word spotting" concept from handwriting recognition community where similar words are searched based on query image or query keyword. The idea of "word spotting" get the edge over conventional recognition procedures where recognition is not easy always and the objective where is to look for the patterns. Word Spotting has been a wide area of interest because of its ability to spot a specified text without being explicitly recognized.

This paper presents an HMM based approach for spotting dates (day, month and year) of different formats (like 29/04/2014 or 29$^{th}$ April 2014 or April 29, 2014 or 4/29/14 or 29-4-2014 or 29.04.14 etc.) from image/video frames. Our date retrieval approach searches the date patterns in individual text lines of a video frame. First, different components of a date such as textual month, numeral, and punctuation are found and this information is used to extract the date patterns. Given a color text line image, first we apply an efficient gray image conversion to enhance the text information. Gradient and wavelet based transformation are combined to extract the enhanced gray image. Next, Pyramid Histogram of Oriented Gradient (PHOG) feature has been extracted from gray image and binary images for date-spotting framework. To boost the performance an efficient MLP (Multi-Layer Perceptron) based Tandem feature is proposed. For spotting purpose, characters based HMMs have been constructed. Finally, a shape coding based scheme is used to combine the similar shape characters in same class during word spotting. Three different date models (A, B and C covering most of the types of date formats) are searched from a line without segmenting characters or words. Our date field extraction method consists of two major tasks namely: month word detectors, and numerical field and date pattern extraction. Hence, the proposed date extraction process from scene images or video frames will be very useful in searching, indexing and interpreting. For examples, portions of scene images and video frames containing date information are shown in Fig.1.

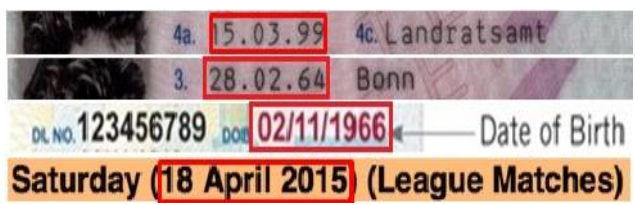

**Fig.1: Example image containing date information**

The contributions of this paper are as follows. First we present a novel feature extraction method for text characters using image enhancement by wavelet and gradient features. We noted that combination of binary image feature and gray image feature improves the HMM-based word spotting performance significantly. Next, a tandem approach using MLP is used to improve the word spotting approach. To further boost the performance, a shape coding based approach is used to combine the similar shape characters in same class during word spotting. The frame work for Latin date-spotting has been tested in multi-script environment to demonstrate the robustness of the proposed system. To the best of our knowledge there exist no other work on date spotting in scene/video images.

The rest of the paper is organized as follows. We will present the related work in Section 2. Section 3 describes the overall framework and provides detailed description of the different steps involved in the framework like preprocessing, feature extraction from binary image and enhanced gray images, MLP-based Tandem modeling for word spotting, and shape coding scheme. Section 4 shows the experiments and the detailed results that have been performed. Finally, conclusion and future works are presented in Section 5.

## 2. Related Work

Though there exist many pieces of work on word spotting [13, 24, 25, 26] in handwritten/printed text lines, not many work exist for text spotting in natural scene image/video frames. Most of the works on scene and video images are on text detection and recognition purpose [27, 40, 41]. It is due to low resolution, blur, background noise, etc. which create hindrance in text recognition in scene and video images. Generally text recognition system in scene images and video frames [27] consists of four steps, (a) text detection which locates text information in the image or video, (b) text extraction which extracts text line from the images, (c) text binarization which separates foreground (text) from background information and (d) text recognition which recognizes text either using existing OCR or developing classifiers. Many pieces of work exist for text detection in both scene and video [28, 29]. To overcome low resolution images, edge analysis is performed to improve text detection [40]. Several methods are there for text binarization by taking text lines or words as input detected by the text detection methods [30, 31, 32]. For scene text recognition, two types of methods exist, using segmentation and without segmentation. Out of these, recognition without segmentation of characters has become popular because segmentation incurs some errors like over-segmentation or under segmentation and thus significant information might be lost due to complex background which may lead to poor recognition rate. For instance, proper character segmentation is not easy in scene text recognition due to segmentation error, a part of a character image may be mis-recognized. Sometimes a part of the letter 'm' could be 'n', or a part of the letter 'w' could be 'v'. In a similar way, the neighbourhood of some characters may be confused with another one, such as 'ri' which can be confused with 'n'. Hence word spotting approaches are preferrable for information retrieval in such noisy images.

**Text Recognition in Scene and Video Images:** For scene text recognition at word level, traditional visual features such as Histogram of Oriented Gradients (HOG) have been proposed by Wang et al. [32]. This method uses lexicons to improve the recognition accuracy. Neuman and Matas [33] proposed a method for scene text recognition by exploiting Maximally Stable Extremal Regions (MSER) and language model. Wang et al. [34] proposed an end-to-end text recognition with convolutional neural networks. They proposed multi-layer networks with unsupervised features learning, which allows a common framework to train highly accurate text detector and character recognizer modules.

Compared to scene text recognition only a few methods are reported for video text recognition [35, 36] in literature because video text recognition is more challenging than scene text recognition. Chen at al. [35] proposed a method using sequential Monte Carlo and error voting for video text recognition. However, this approach is sensitive to thresholds and segmentation of characters. An automatic binarization method for colour text area in video based on convolutional network is proposed by Saidane and Garcia [36]. The performance of the method depends on the number of training samples. In the same way, edge based binarization for video text image is proposed by Zhou et al. [30] to improve the video character recognition rate. This method takes Canny of the input image as input and it proposes modified flood fill algorithm to fill the gap if there is a small gap on contour. This method does not work well for big gaps on character contours. In addition, the method is sensitive to seed points.

**Word Spotting in Scene and Video Images:** Word spotting, as mentioned earlier, is an extensively practiced area of research considering texts in handwritten [4-6, 24, 26] or printed documents [7, 25] even in different scripts [8]. Although, couple of works [9, 10] show some efficient approaches of text detection in scene images based on background invariant features [11], etc. but the problem is still not solved in general. The existing methods work for specific formats of date and also work on specific documents such as bank cheques. Suen et al. [19] proposed an approach for automatic cheque processing i.e. the segmentation and recognition of dates written on bank cheques. Recently, a two-staged classification-based approach has been proposed for date field extraction from handwritten documents [18]. A word-level and component-level classification was done to locate the date components.

## 3. Proposed framework

In this paper, an HMM based approach has been proposed for spotting dates in a given video frame/scene image. A flowchart of our proposed framework is shown in Fig.2. For this purpose, the text lines segmented from the scene image/video frames are binarized using an efficient Bayesian classifier based binarization approach [17]. Next, features are extracted from these binary images and enhanced gray images to feed into HMM-based date-spotting system. Next, a tandem approach using Multi-Layer Perceptron is used to improve the word spotting approach. To further boost the performance, a shape coding based approach is used to combine the similar shape characters in same class during word spotting. Details of these modules are discussed in following subsections. Date components and their modeling are also explained for spotting purpose.

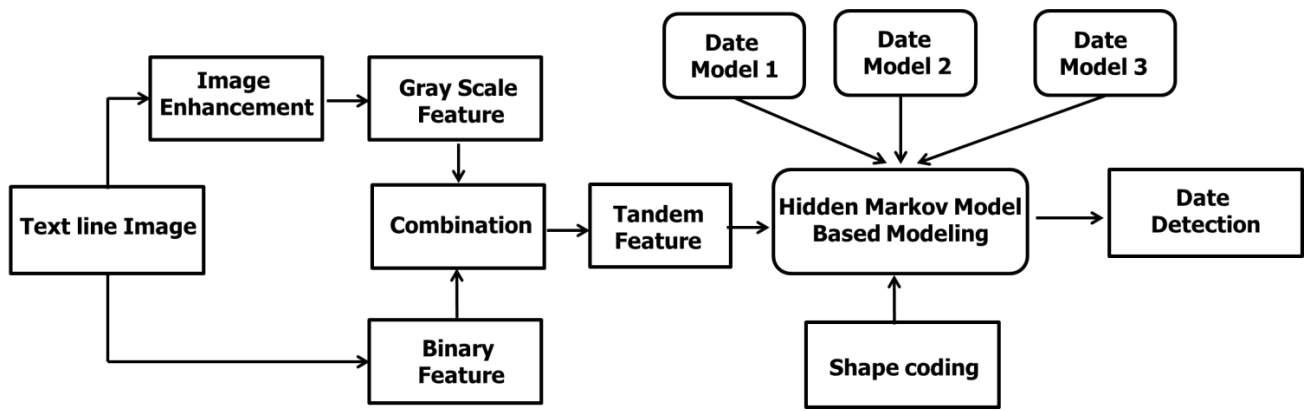

**Fig.2: Flowchart of the proposed date-spotting framework.**

### 3.1. Date Components

There may be four types of components in a date field of any video frames (numeral, textual month, punctuation and contraction). Date patterns can be classified into two categories on the basis of different component types:
- Numeric dates; e.g. 22/03/2006, 10-11-15, 6.5.1994 etc.
- Semi-numeric dates; e.g. 22nd March, 2006, March 9, 2010, Oct 2$^{nd}$ 1980 etc.

**Numeric dates:** A date field consisting of numerals and punctuations (examples of numeric dates are 06/05/15, 6/5/2015, 6-5-15 etc.) are considered to be numeric date fields. It has been observed that the total number of components in a numeric date field may vary from 6 to 10 (If a date is written as 3/8/10 then the number of components of this date is 6. If a date is written as 03/08/2010 then the number of components will be 10). The following date regular expression represents the valid numeric date formats:

$$(d|dd) \ (-./) \ (d|dd) \ (-./) \ (dd|dddd)$$

where, 'd' represents a digit (0-9). A date has three parts or fields: date field (1-31), month field (1-12) and year field (00-99 or 0000-9999). A complete numeric date field consists of a single digit or double digit date information (d | dd), single digit or double digit month information (d | dd) and double digit or four digit year information (dd | dddd) with two punctuation marks to separate day, month and year information.

**Semi-numeric date:** Date fields that consist of textual month (examples are January, JANUARY, February, FEBRUARY etc.) and non-month word block identification (examples are Jan, Feb, Mar, Apr, Aug, Oct, Dec), digits (like 0-9) and contraction (examples are st, nd, rd, th) are considered as semi-numeric dates. Some examples of semi-numeric dates are October 3$^{rd}$ 1990, May 9, 2010, 31$^{st}$ Dec, 2011. For semi-numeric date field extraction, we search for the following regular expressions:

$$(md \ | \ mdd) \ (-.,) \ (dd \ | \ dddd) \quad \text{and} \quad (d \ | \ dd) \ (contraction) \ (month) \ (-.,) \ (dd \ | \ dddd)$$

where, 'm' represents a month field and 'd' represents digit field. There are two types of sequence for semi-numeric date fields depending on the position of the textual month field (examples of date with month information in the middle position: 15th August, 2015 and month information in the starting position: August 15, 2015).

Based on the above classification, we have used three different models, to fit all the possible date patterns found in the dataset. Three models are as follows:
- Model A; e.g. 06/05/1994.
- Model B; e.g.1$^{st}$ May 1986.
- Model C; e.g. Oct 3$^{rd}$ 95.

**Model A:** It has been used to spot the dates with numbers that are separated by punctuations (slash **'/'**, hyphen **'-'** and period **'.'**) and do not include month names, some more formats are mm**-**dd**-**yyyy, d**.**m**.**yy, d/m, d**.**mm**.**yy, etc. where 'd' denotes date field, 'm' denotes month field and 'y' denotes year field, all represented by digits (0-9) and separated by punctuations like '.' , '/' , '-' (Fig.3a). Some examples of model A are 05/06/10, 5/6/2010, 5-6-10 etc.Whereas, month names have been taken care of using Model B and Model C.

**Model B:** It has been used to model dates with the date field containing numerals with their respective suffixes (examples are st, nd, rd, th) if present followed by the months name written in their full (e.g. "December") as well as abbreviated forms (e.g. "Dec") along with a year format, if present (Fig.3b). Some examples of model B are 2nd March, 2015, $23^{rd}$ Dec, $23^{rd}$ Jan 97, 01 OCT 07 etc.).

**Model C:** It accounts for the dates with the month name written in their full (e.g. January) as well as abbreviated forms (e.g., Jan), followed by the date field denoted by digits (0-9), suffixes (examples are st, nd, rd, th) if present and year in different formats (Fig.3c). Some examples of model B are November 7, 1994, April $29^{th}$ 2014, Sept 15, Jan. 15, 2014 etc.).

Most of the variations in month names and style have been taken in account in the work to spot all possible date formats that we encounter during the training and testing process.

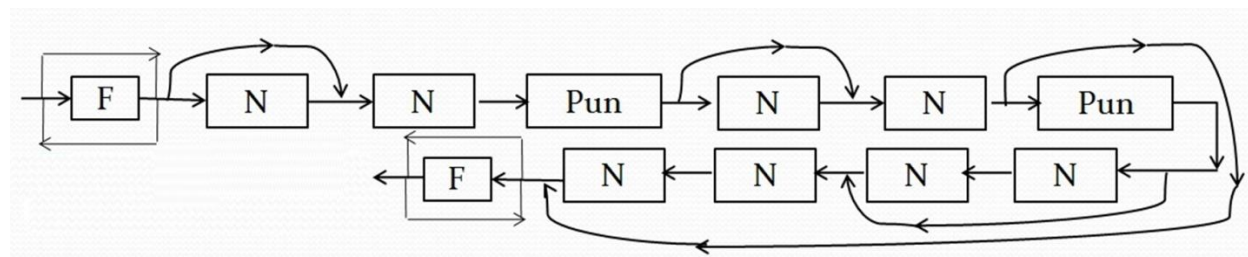

(a)

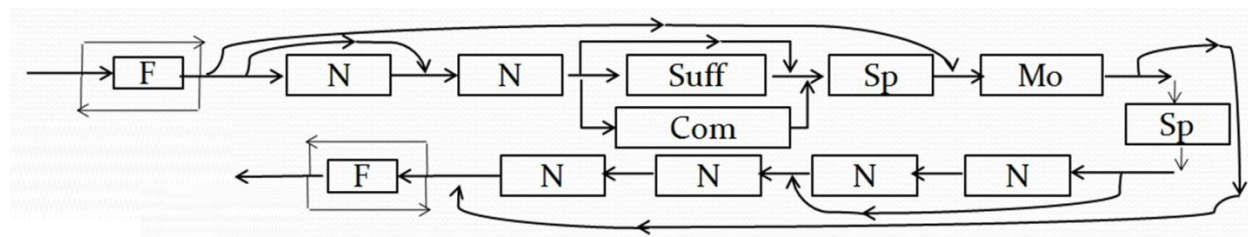

(b)

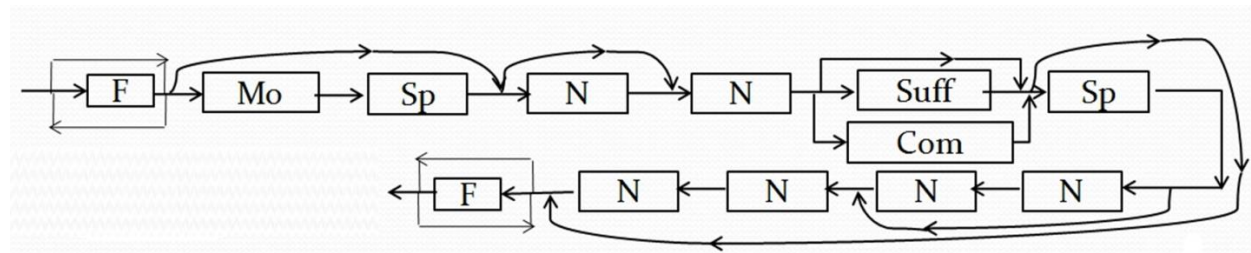

(c)

**Fig.3: Graphical representations of three different date models are shown. (a) Numeric date model, e.g. 06/05/1994 (b) Semi-numeric date model with month name in the middle, e.g. 1st May 1986 (c) Semi-numeric date model with month name at the beginning, e.g. Oct 3rd 95. Here the acronyms are borrowed from Fig.4. "Sp" and "Com" represent the "space" and the "comma" respectively.**

## 3.2. Hidden Markov Model (HMM) based Date Modeling

### 3.2.1 Hidden Markov Model

Hidden Markov Model (HMM) based approach is used in our system for spotting date patterns to locate the date field in video frames or natural scenes. The simplest model of character HMM which consists of J hidden states ($S_1$, $S_2$ ... $S_J$) in a linear topology as an observation **O** where i[th] observation ($O_i$) represents an n-dimensional feature vector **x** modeled using a Gaussian Mixture Model (GMM) with probability $P_{S_j}(x)$, 1<j<J given by

$$P_{S_j}(x) = \sum_{k=1}^{G} W_{jk} N(x|\mu_{jk}, \Sigma_{jk})$$
(1)

Where G is the number of Gaussians and *N* refers to a multivariate Gaussian distribution with mean **$\mu_{jk}$**, covariance matrix **$\Sigma_{jk}$** and probability **$W_{jk}$** for k[th] Gaussian in state j.

For spotting purpose, we have used three different models constructed by the character HMMs [13], to fit all the possible date patterns found in the dataset. Some date formats are like, dd/mm/yyyy, dd/mm/yy, dd/mm, dd[th] month yyyy, month dd yyyy, etc (dd, mm, yy are shorthand representations of day, month and year digits respectively). Some supporting date-model units (See Fig.4) for numbers, punctuations, month names, years, ordinal number suffixes etc. have been created. Number model consists of all the digits from '0' to '9' in parallel. Punctuation model comprises of slash '/', hyphen '-' and period '.' which can usually be seen in regular dates. The ordinal number suffixes (i.e. 'st', 'nd', 'rd', 'th') are used often in dates (e.g. 29th Jan). The months are written in their full (e.g. 'January') as well as abbreviated forms (e.g. 'Jan') in many places.

These units are next used in date keyword model in a hierarchical manner. The generic date formats mentioned above were considered as keywords instead of specific dates. Three complete date keyword models are shown in Fig.3. For spotting dates, we have used the previously built (as shown in Fig.4) supporting models to construct symbolic representations.

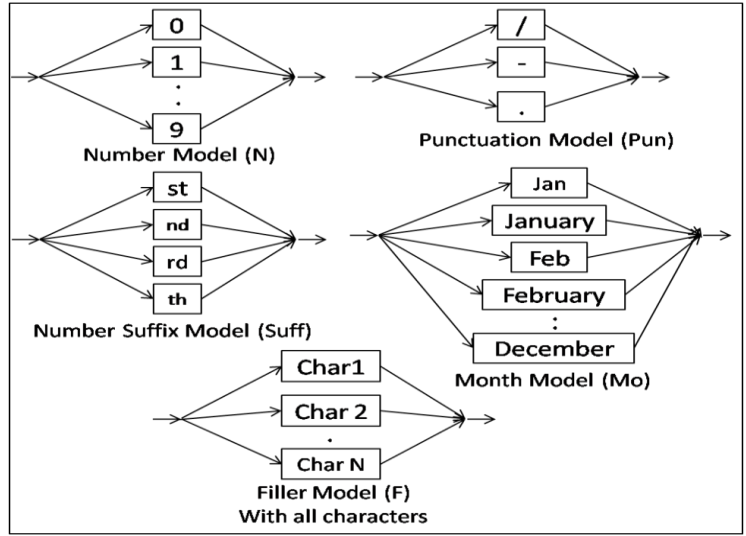

**Fig.4: Supporting models (Acronyms of the supporting models in brackets have been used in the date models)**

### 3.2.2 Text Line Scoring

Our video date spotting mechanism depends on the scoring of text images containing date-field(X) for the keyword (W), which gives a positive value for the occurrence of that particular keyword when the score value exceeds a certain threshold in that text line. The posterior probability $P(W_j|X_{a,b})$ trained on keyword models assigns the score value to the text line image, where $X_{a,b}$ denotes the part of the text line containing the keyword and a,b corresponds to the start and end position of the keyword. Applying Bayes' rule we get

$$\log p(W|X_{a,b}) = \log p(X_{a,b}|W) + \log p(W) - \log p(X_{a,b}) \qquad (2)$$

Where $\log p(X_{a,b}|W)$ represents the keyword text line model, the $\log p(W)$ term is ignored due to equal probability and the remaining part is modeled with Filler text line model. Now, to find the positions of a, b for the keyword we use the log-likelihood:

$$\log p(X_{a,b}|W) = \log p(X_{a,b}|K) \qquad (3)$$

This helps us to get the general conformance of the text image to the trained character models given by $\log p(X_{a,b}) = \log p(X_{a,b}|F)$, where the term $\log p(X_{a,b})$ represents the unconstrained filler model F. The length of the word is used for normalization between the log-likelihood value of keyword model and filler model to get the final text line score as:

$$Score(X, W) = \frac{[\log p(Xa, b \mid K) - \log p(Xa, b \mid F)]}{b - a} \qquad (4)$$

## 3.3. Feature Extraction

### 3.3.1. Text enhancement

Roy et al. [17, 46] proposed integration of three domains (RGB, wavelet and Gradient) to enhance the word image in scene image/video for recognition. It was noted that the pixel value of text component might be low in one sub-band but high in another sub-band. The integration of sub-bands of different domains exploits this information for enhancing text in the word image. It was found that the probability of a pixel being classified as text is high compared to a non-text pixel getting a low probability value. In our approach we have used combination of RGB, wavelet and Gradient-based enhancement approach [17] as pre-processing to extract the features from text lines.

To enhance the text pixels in gray image, for each line image (I), the method decomposes it into R, G, B sub-bands in the color domain, LH (Horizontal), HL (Vertical), HH (Diagonal) in the wavelet domain and Horizontal, Vertical, Diagonal in the gradient domain. Then for each set of sub-bands, three sub-bands are combined to obtain three combined images of the respective domains, namely, RGB-L, Wavelet-L and Gradient-L to improve the fine details at the edge pixel. The linear combination combines three pixels in the respective three sub-band images by adding three values of each corresponding pixel. Due to complex background, there are chances of introducing noise pixels by the linear combination operation. To take care of this issue, a histogram analysis using a sliding window technique over each linearly combined image was used to select the dominant pixels which are considered as text pixels. A smoothing approach [17] was used to remove the high contrast noisy pixels. Finally, we integrate the three smoothed images of the three respective domains by assigning a weight for each pixel. Determination of weight was performed by computing the variance in a neighborhood to select the high frequency coefficients [20]. Finally, a fusion scheme was used for the high-frequency bands as below. Here D refers to smooth image of RGB, Wavelet and Gradient respectively and σ denotes the variance of the local window surrounding the pixel [17].

$$D_F = \begin{cases} D_{RGB}, & \text{if } \sigma_{RGB}(I) = \text{Max}(\sigma_{RGB}(I), \sigma_W(I), \sigma_G(I)) \\ D_W, & \text{if } \sigma_W(I) = \text{Max}(\sigma_{RGB}(I), \sigma_W(I), \sigma_G(I)) \\ D_G, & \text{if } \sigma_G(I) = \text{Max}(\sigma_{RGB}(I), \sigma_W(I), \sigma_G(I)) \end{cases} \quad (5)$$

### 3.3.2. Feature extraction from Gray and Binary image

We use Pyramid Histogram of Oriented Gradient (PHOG) [3] feature extraction approach from enhanced gray scale image. We also considered binary version of the text-line image to obtain the PHOG feature. These are detailed in following subsections.

**Pyramid Histogram of Oriented Gradient (PHOG) feature:** PHOG [3] is the spatial shape descriptor which gives the feature of the image by spatial layout and local shape, comprising of gradient orientation at each pyramid resolution level. To extract the feature from each sliding window, we have divided it into cells at several pyramid level. The grid has $4^N$ individual cells at $N$ resolution level (i.e. $N=0, 1, 2..$). Histogram of gradient orientation of each pixel is calculated from these individual cells and is quantized into $L$ bins. Each bin indicates a particular octant in the angular radian space.

The concatenation of all feature vectors at each pyramid resolution level provides the final PHOG descriptor. L-vector at level zero represents the L-bins of the histogram at that level. At any individual level, it has $Lx4^N$ dimensional feature vector where N is the pyramid resolution level (i.e. N=0, 1, 2….). So, the final PHOG descriptor consists of $L \times \sum_{N=0}^{N=K} 4^N$ dimensional feature vector, where $K$ is the limiting pyramid level. In our implementation, we have limited the level (N) to 2 and we considered 8 bins (360º/45º) of angular information. So we obtained (1×8) + (4×8) + (16×8) = (8+32+128) = 168 dimensional feature vector for individual sliding window position.

**Text Binarization using Bayesian Classifier:** As mentioned earlier the probability of pixels in RGB-smooth, wavelet-smooth and gradient-smooth being classified as text has high values compared to non-text pixels. A simple probability calculation that if a pixel in RGB, Wavelet and Gradient images gets high value (towards 1) at the same location of the three images then the probability of that pixel to be a text pixel is considered as high. Similarly, if a pixel in RGB, Wavelet, and Gradient at the same location of the images gets a low value (towards 0) then the probability of that pixel is considered to be a non-text pixel as high. Due to this property, a Bayesian classifier for binarization was proposed in [17].

In Bayesian framework, the number of classified text pixels and number of classified non-text pixels are considered as a priori probability of text pixel class and non-text pixel class, which are denoted as P(CTC) and P(NCTC), respectively. P(f(x, y)|TC) denotes the conditional probability of a pixel (x, y) for a given Text Class (TC) which is average of RGB-Smooth, Wavelet-Smooth and Gradient-Smooth and P(f(x, y)|NTC) denotes Non-Text Class (NTC) which is obtained by taking average of complement of RGB-Smooth, Wavelet-Smooth and Gradient-Smooth images. Hence, the conditional probabilities and priori probabilities are substituted in Bayesian framework as given below to obtain posterior probability matrix [17].

$$P(TC|f(x,y)) = \frac{P(f(x,y)|TC)*P(CTC)}{P(f(x,y)|TC)*P(CTC)+ P(f(x,y)|NTC)*P(NCTC)} \qquad (6)$$

Then the final binary image (B(x, y)) is obtained with the condition given in following equation on posterior probability matrix.

$$B(x,y) = \begin{cases} 1 & \text{if } P(TC|f(x,y)) \geq \gamma; \\ 0 & \text{Otherwise.} \end{cases} \qquad (7)$$

where γ is the threshold parameter which is set to 0.05 [17]. An example of text line and its corresponding binary image are shown in Fig. 5.

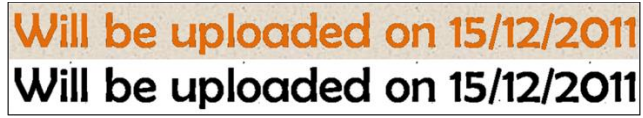

**Fig.5: Binary image using text binarization method [17]**

### 3.3.3. Tandem Approach Combining Gray and Binary Image Features

We present an efficient Tandem-based feature for Date spotting in our system. Tandem modeling [37] proposed to combine the discriminative power of the Artificial Neural Networks with the sequence modeling ability of the HMMs in speech recognition. The positive effect of the combined feature is that the Multi-Layer Perceptron (MLP) performs a non-linear feature transformation into a space that is explicitly oriented for discriminability of characters/states. The transformed feature leads to improved discrimination by the GMM which describes the output space associated with each HMM state. The advantage of the tandem approach is that it is robust to noise.

We present the MLP-HMM tandem systems for our word spotting purpose in Fig. 6. At first, binary and gray image features each having 168 dimensions is combined to make 336 dimensional feature. Next, MLP based discriminative features are integrated into the HMM framework to form the tandem MLP-HMM system. In contrast to the likelihood of a feature vector $x_t$, given an arbitrary state $s_i$, MLPs produce state posterior probability $P(s_i | x_t)$. Training the MLP requires each observation at time step *t* in the training data to be aligned to a character label in the word. However, the class (e.g., HMM-state) labels are usually not available. To do so, a previously trained GMM-HMM is applied to the training data in the forced alignment mode. Next, the MLP is trained on the labeled observations in a frame-based approach. The trained MLP is used to compute the posterior distribution over the character labels for each observation. We considered 66 character levels in our method. To perform the sequence modeling in a tandem HMM approach, the posterior estimates are considered as observations to train a new HMM (GMM-HMM). Generation of Tandem feature is illustrated in Fig.6. The output posterior probabilities are decorrelated by a dimensionality reduction algorithm. We applied Principal Component Analysis (PCA) [38] to the posterior probabilities of the MLPs. It is done to reduce the dimensionality and to orthogonalize the feature vectors. Next, the features were normalized by mean and variance. Finally, these reduced feature vectors were concatenated with the baseline 336 dimensional HMM features.

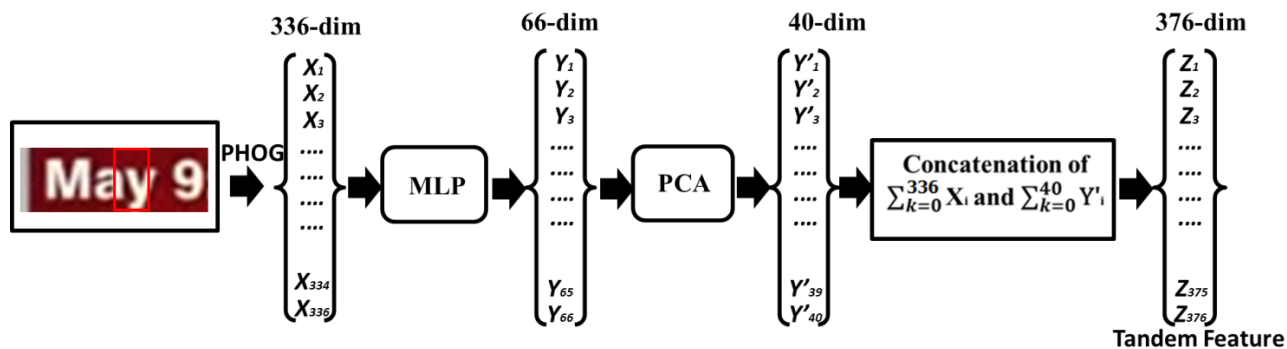

**Fig.6: Combination of binary and gray images features in a Tandem model**

### 3.3.4 Shape Coding based Word Spotting

Shape coding based text encoding approach has been used efficiently in printed documents [21, 22]. These approaches annotate character images by a set of predefined codes. Nakayama [22] annotated character images by seven shape codes for word content detection. Lu et al. [21] proposed a set of topological codes based on shape features including character ascenders/descenders, holes, water reservoirs information, etc. to retrieve document images. Usually, these shape codes were manually designed to perform text document searching.

Inspired with this idea, the proposed date spotting approach uses HMM-based date spotting using shape coding. Similar-shaped text characters are grouped together and they are encoded according to [21]. The word spotting approach is next modeled using HMM by modified shape coding based character models. Table I shows the shape coding mapping used in our approach. All characters are next encoded and modeled using Tandem approach. Finally, date patterns are searched by this shape coded character HMM approach.

**Table I: Shape coding representation used in [21].**

| Character | Shape Code | Character | Shape Code | Character | Shape Code | Character | Shape Code |
|---|---|---|---|---|---|---|---|
| A | A | Q | Q | g | g | v | v |
| B | B | R | R | h | I | w | vv |
| C | C | S | S | i | i | x | uc |
| D | O | T | l | j | j | y | y |
| E | E | U | ll | k | lc | z | z |
| F | F | V | ll | l | l | 0 | O |
| G | C | W | ll | m | m | 1 | I |
| H | ll | X | uc | n | m | 2 | 2 |
| I | l | Y | ll | o | O | 3 | 3 |
| J | l | Z | Z | p | no | 4 | O |
| K | lc | a | a | q | on | 5 | 5 |
| L | l | b | lo | r | r | 6 | 6 |
| M | ll | c | c | s | s | 7 | l |
| N | ll | d | ol | t | lc | 8 | B |
| O | O | e | e | u | v | 9 | 9 |
| P | P | f | f | | | | |

## 4. Experiment Results & Discussions

As there exist no such dataset of scene/video with text lines containing different date patterns, we have collected our own dataset to measure the performance. For training and testing our system we divided the dataset into two sets of data in which majority of them are captured from various videos (such as videos of news bulletin, events, match/exam schedule etc.) having dates and remaining from scene images.

## 4.1. Dataset

For our experiment we collected a total dataset of 10,053 images out of which 6,886 images are used as training, 1519 images are considered as validation data and rest 1648 are used as testing data. The data is collected from news channel, youtube video, web services, etc. The ground truths of the dataset are generated manually. Dataset is made available online for further research in this direction [39]. The training samples which were selected randomly from the dataset consist of all possible valid date fields including numeric (consisting of digits from 0-9), alphabetic characters (consisting of alphabets from a-z and A-Z used for months and suffixes), punctuations (e.g. slash '/', hyphen '-' and period '.'), etc. of different fonts and sizes to built an efficient system that could spot a large variety of date patterns and the testing dataset consists of images having date field of all different patterns to check the effectiveness of our system. Fig.7. provides few examples from our dataset.

| Type | Training | Validation | Testing |
|---|---|---|---|
| Scene Image | | | |
| Video Image | | | |

**Fig.7: Some examples of scene and video images and corresponding text lines considered in our experiment.**

We have measured the performance of our word spotting system using precision, recall and mean average precision (MAP). The precision and recall are defined as follows.

$$Precision = \frac{TP}{TP+FN} \quad Recall = \frac{TP}{TP+FP} \tag{8}$$

Where, TP is true positive, FN is false negative and FP is false positive. MAP value is evaluated by the area under the curve of recall and precision. We have checked the performance using both global and local threshold where, global threshold correspond to each tuning the threshold value over all the images in the dataset and local threshold indicates optimizing the threshold value for a particular testing image to get optimum performance.

## 4.2. Date Retrieval Performance using Proposed Approach

The training is performed with the popular HTK toolkit [14]. We used continuous density HMMs with diagonal covariance matrices. Parameter evaluation in HMM is detailed in Section 4.4.

### 4.2.1. Performance using Binary Image Features

Experiments with three different date models are performed in our dataset to evaluate the performance. The date models are leaned by the grammar described in Fig.3. For the experiment we considered 32 as Gaussian mixture and 8 as state number for HMMs. While testing, the spotting results are then checked with the ground-truth of the test data to find the accuracy. The log-likelihood scores are thresholded by a global threshold value [24] to distinguish between correct and incorrect spotting.

For qualitative analysis, some of the keywords that have been spotted by our system along with incorrect spotting have been shown in Table II. Note that, proposed date models are efficient in date pattern detection in complex line images with different background of color and texture. Some of the spotting results are incorrect due to poor resolution and similar date-line text patterns. It may also happen that the punctuation marks are not detected properly due to low resolution and blur of images and hence error occurred in number strings due to appearance of similar date pattern.

Table II: Qualitative results with correct as well as incorrect spotting

| Models | Example images | Status |
|---|---|---|
| Model A | John R. Doe  Pay Period 06/02/06 | ✓ |
| | DL NO. 123456789 DOB 02/11/1966 | ✓ |
| | 04/18/2014 14:45 | ✓ |
| | $300.00 | Monday | ✗ |
| Model B | INDIA | ✗ |
| | 16 MAY, 6 AM ONWARDS | ✓ |
| | | Monday 3 March 2014 | ✓ |
| Model C | ISRO on Sept 14, | ✓ |
| | Sept 1, 2015 (Steam®) | ✓ |
| | May 31: Audio recording released | ✓ |

Three date models (keywords) have been tested separately for quantitative analysis. The precision-recall curve is shown for each of the three models (Model A, B and C discussed in section 2.3) below. We have noted that from Fig.8 Date format of Model B (e.g. 2nd March, 2015, etc.) shows better performance among these three models.

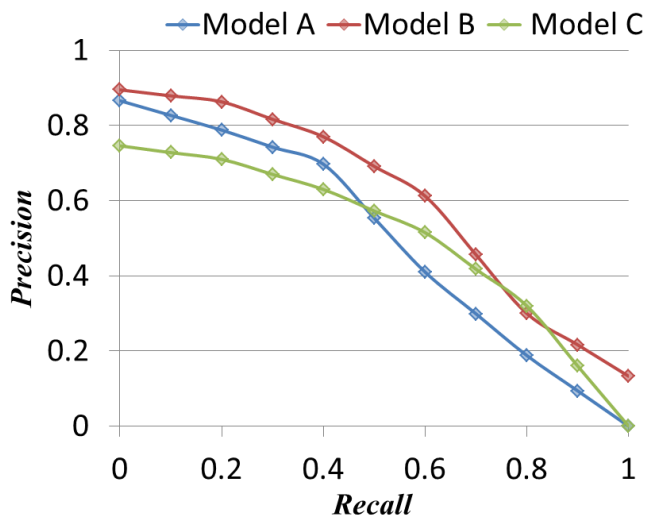

**Fig.8: Comparative study of data spotting performance using our proposed approaches using Model A, Model B and Model C respectively using binary image feature.**

**4.2.2. Performance using Gray Image Features**

After enhancing the original text line images using our Bayesian framework by information combining RGB-Smooth, Wavelet-Smooth and Gradient-Smooth images, the gray images are tested for date spotting. We have noted that gray image information improves the spotting performance. Fig. 9 shows the precision-recall curves obtained from Binary and gray image text lines.

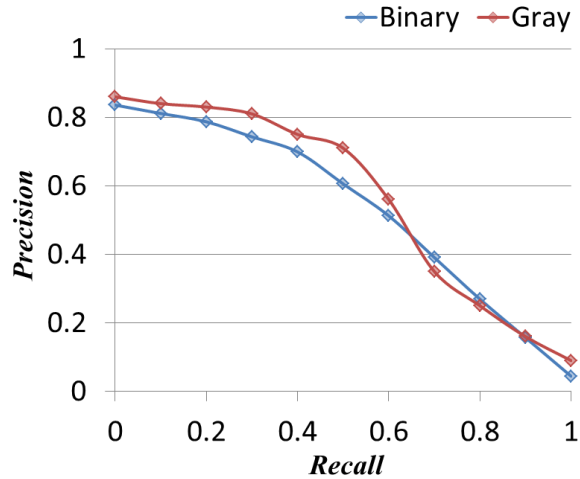

**Fig.9: Comparative study of data spotting performance using binary image feature and gray scale image feature where we considered an average of the three models.**

**4.2.3. Performance using Tandem approach combining binary and gray scale feature**

Binary and gray image features are combined next to check the date spotting performance. It was noted that combination of binary and gray image features improves the individual performance. Tandem-based combination approach is next considered for date spotting. We noted that precision-recall curve is improved after using Tandem-based features. Fig.10 shows data spotting performance by combination of binary image feature and gray scale image feature using tandem feature. The MLP-based Tandem-feature captures efficiently the binary and gray information from low resolution video frame. Different dimension of features were evaluated in PCA and finally we considered 40-dimension according to experiment results.

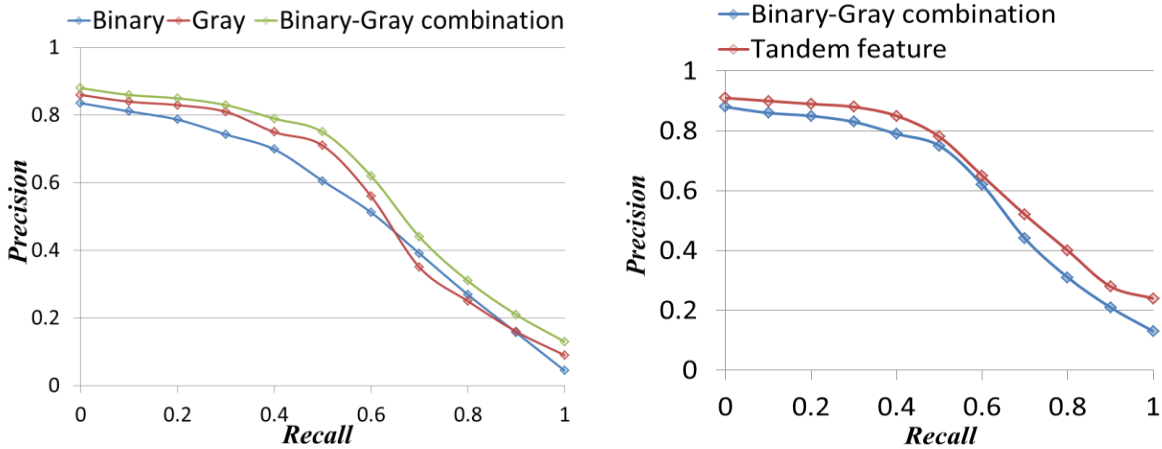

**Fig.10: Comparative study of data spotting performance using combination of binary image feature and gray scale image feature (a) without using Tandem and (b) using Tandem feature.**

### 4.2.4. Date Spotting using Shape Coding

We have integrated next shape coding based scheme in date spotting by Tandem-HMM. Similar character shapes are combined to reduce the character confusion and hence spotting improvement. Fig.11 shows the date detection performance with and without using shape coding. It is due to merging of similar shaped characters together. The *F measures* and corresponding precision and recall values are shown in Table III. A comparative study of qualitative results is shown in Fig.12. Both local and global MAP values of with and without shape coding based approach along with different types of feature is provided in Table IV.

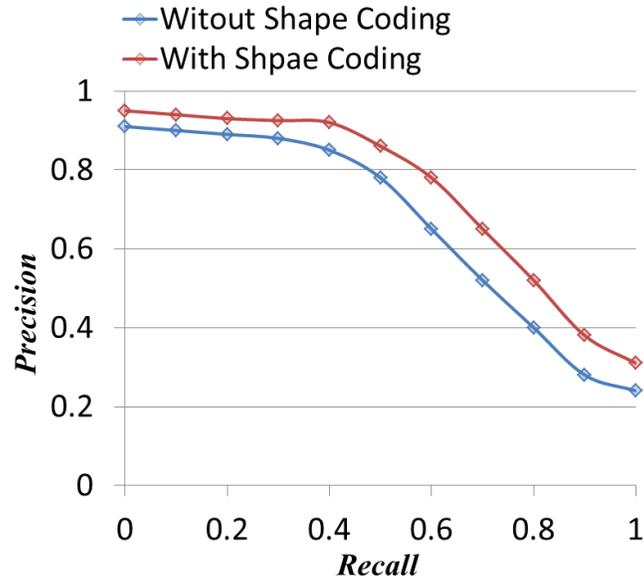

**Fig.11: Comparative study of data spotting performance using traditional approach and shape coding based character encoding.**

**Table III: Comparative study between with and without shape coding based method**

| Method | Precision | Recall | F – Measure |
|---|---|---|---|
| Without shape coding | 74.58 | 75.12 | 74.84 |
| With shape coding | 82.58 | 83.21 | 82.89 |

**Table IV: Comparative study using different approach and feature type.**

| Method | | MAP Value | |
|---|---|---|---|
| Approach | Feature Type | Local | Global |
| Without Shape Coding | Binary Feature | 78.42 | 64.19 |
| | Gray-scale Feature | 89.12 | 65.01 |
| | Binary + Gray-scale Feature | 82.45 | 67.48 |
| | Binary + Gray-scale Feature + Tandem | 84.18 | 69.14 |
| With Shape Coding | Binary + Gray-scale Feature + Tandem | 89.47 | 76.47 |

| Example Images | (i) | (ii) | (iii) | (iv) | (v) |
|---|---|---|---|---|---|
| Date & Time  01/20/2011;14:06 | ✓ | ✗ | ✓ | ✓ | ✓ |
| RECEIVED  AUG 28 2014 | ✗ | ✓ | ✓ | ✓ | ✓ |
| 04/26/2010 | ✗ | ✗ | ✗ | ✗ | ✓ |
| Start time . : 08/19/2009 18:00:00 | ✗ | ✗ | ✓ | ✓ | ✓ |
| 2015/03/13 14:38:12 | ✓ | ✓ | ✓ | ✓ | ✓ |
| 09-13-2011 03:41:16 PM | ✗ | ✗ | ✗ | ✗ | ✓ |
| 2/20/2015 | ✗ | ✓ | ✓ | ✓ | ✓ |
| MON - DEC 07, 2015 | ✓ | ✓ | ✓ | ✓ | ✓ |
| 27-08-91 TUE | ✗ | ✗ | ✗ | ✓ | ✓ |
| PM 7:32 OCT.21.2012 | ✗ | ✓ | ✓ | ✓ | ✓ |

**Fig.12: Example showing comparative study of date spotting performance using (i) Binary Feature (ii) Grayscale feature (iii) Binary-Gray combination (iv) Tandem Feature (v) Shape coding**.

## 4.4. Parameter Evaluation

We considered continuous density HMMs with diagonal covariance matrices of GMMs in each state. The window slides with 50% overlapping in each position. Experiments were carried out by varying the number of states from 2 to 8. The best state number was found to be 6. A number of Gaussian mixtures were tested on validation data. The numbers of Gaussian distributions were considered from 1 to 128 increasing with a step of power of 2. Word spotting performance with different Gaussian numbers is detailed in Fig.13(a). It was observed that with 32 Gaussian mixtures, all features provide the best results. Fig.13(b) illustrates the performance with varying the state number on word spotting experiment.

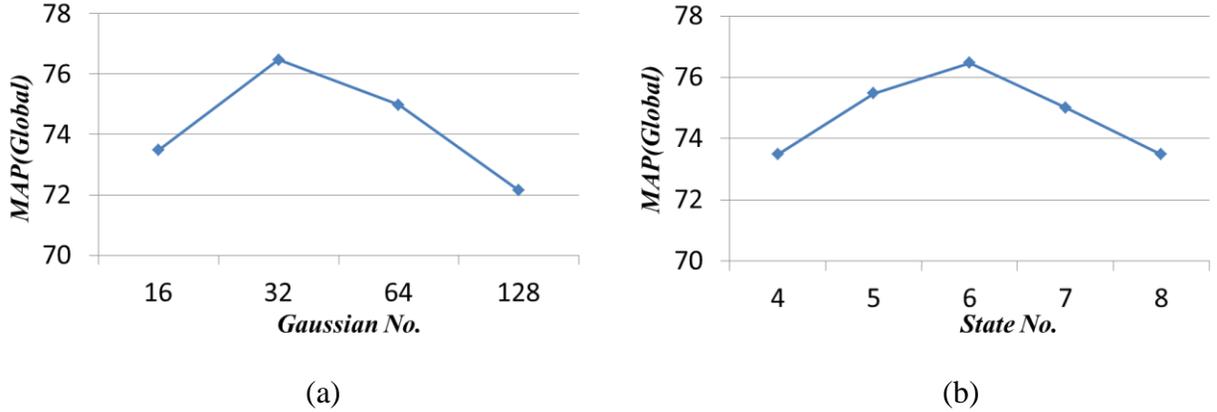

**Fig.13: Word spotting performance evaluation using different (a) Gaussian number and (b) State number**.

## 4.5 Comparative Study

A comparison between proposed HMM-based system and the Google Tesseract-OCR [12] has been done to understand the effectiveness of our method. Video frames were fed to Tesseract-OCR. It is found that it performed well in some cases but failed considerably for images with low resolution, blur texts, complex background, etc. Whereas our proposed system was able to detect almost all varieties of date formats. Some examples where Tesseract-OCR failed but our system succeeded are given in Fig.14. For quantitative analysis, the percentages of correct spotting with our method using all three models and with Tesseract-OCR are shown in Table V.

We have compared the performance with Marti-Bunke feature [2] which is used extensively for Latin script recognition [47]. This profile based feature consists of nine features computed from foreground pixels in each image column. Out of these, three features are used to capture the fraction of foreground pixels, the centre of gravity and the second order moment and remaining six features comprise of the position of the upper and lower profile, the number of foreground to background pixel transitions, the fraction of foreground pixels between the upper and lower profiles and the gradient of the upper and lower profile with respect to the previous column. The performance of date detection in our dataset using this feature is provided in Table V. Since, the Marti-Bunke feature [2] captures profile information for character modelling; the performance in our scene/video dataset is poor. It is due to mainly low resolution of such images.

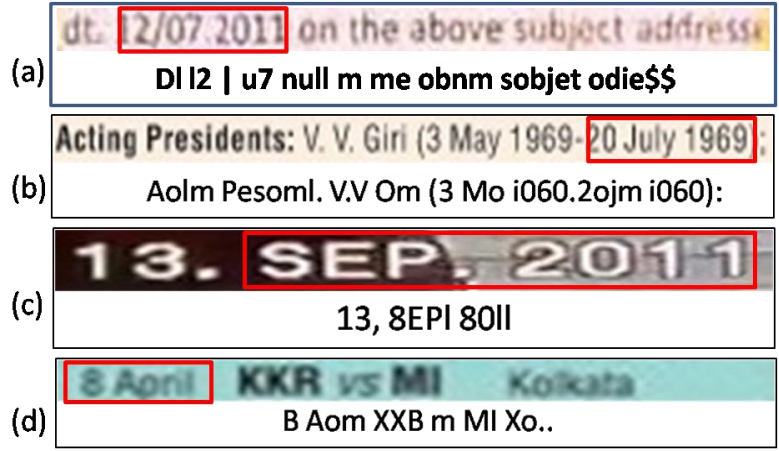

**Fig.14: Wrong recognition (recognition results are shown below each text line image) by Tesseract-OCR & spotting result of our proposed system (shown by red boxes)**

**Table V: Comparison of date detection with proposed method, Marti-Bunke [2] feature and Tesseract-OCR (Global MAP Value).**

| Proposed method | Tesseract-OCR | Marti-Bunke [2] |
|---|---|---|
| 76.47% | 31.57% | 48.12% |

## 4.6 Error Analysis

We found out few images which weren't properly recognized by our binarization method due to very low resolution, blur and complex color transition. Some examples of improper binarization are shown in Fig.15.

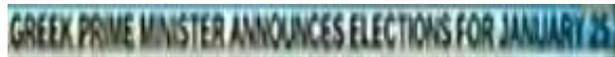
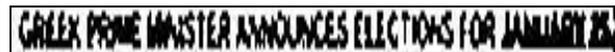

(a)

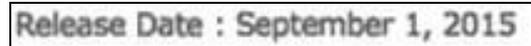
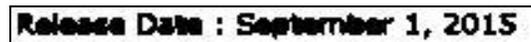

(b)

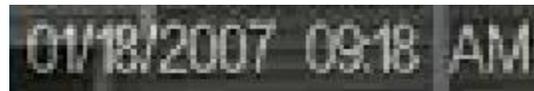
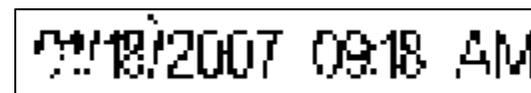

(c)

**Fig.15: Poorly binarized image where our date spotting system failed.**

In videos, sometimes we encounter different fonts than the conventional ones. Because of cursive/stylish fonts, it may be difficult to spot any text of such font in video frames. For getting precise spotting even with various fonts, single character of different fonts should be embedded inside the character HMMs by training all possible fonts which seems to be tedious as there is a large number of fonts used by television channels, video editors, etc. An example of incorrect recognition is shown in fig.16.

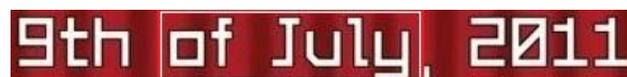

**Fig.16: Wrong spotting due to font variation (The example is tested to find the keyword of pattern 'dd Month' while it is recognized as '01 July' and this is because of shape similarity)**

In model 'A', the date format having only numerals, we encountered an error in which we saw that our proposes system could not recognize the day and month field when they exceed their respective value of 31 and 12 and spots it as a valid date format. An example of which is shown in Fig.17.

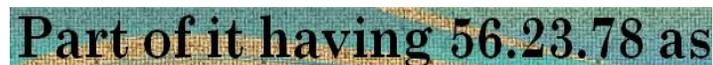

**Fig.17: Wrong spotting of date due to similar date format but day, month exceeding their limiting values.**

## 4.7. Experiment with Synthetic Noise

To test the robustness of our approach we have tested our system with images added with synthetic noises. The text lines images are degraded with Gaussian noise of different noise levels (10%, 20% and 30%). Some examples of synthetic noise added image are shown in Fig.18. Here, the word images are added with 10% Gaussian noise. Quantitative results with noisy images obtained by 10%, 20%, and 30% Gaussian noise are shown in Fig.19. We noted that, date spotting performance in 10% and 20% added Gaussian noised were competitive. Whereas Tesseract-OCR performed very poor in such images.

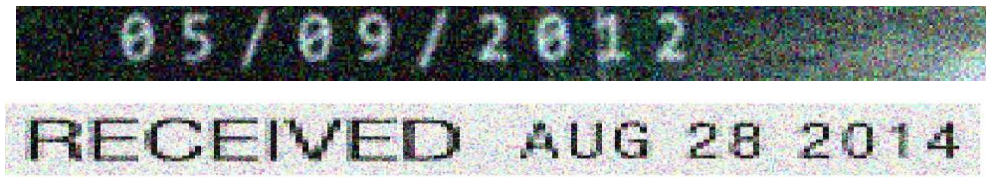

**Fig.18: Images with 10% Gaussian noise added**

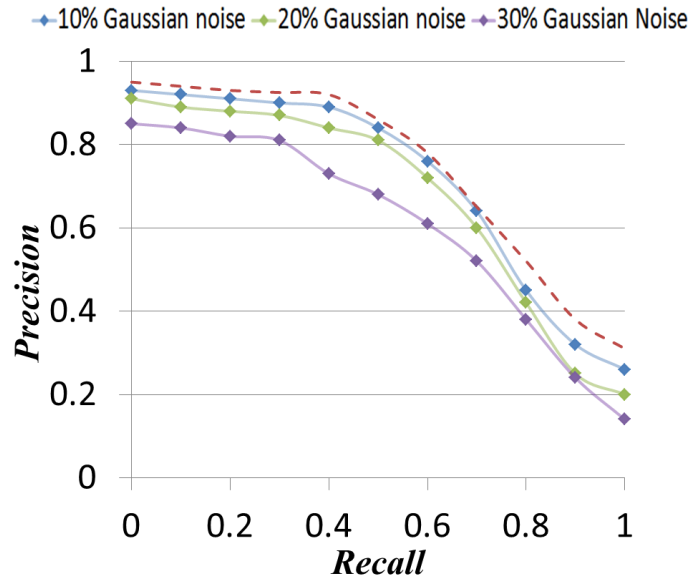

**Fig.19. Date spotting performance using noisy data. Dotted line denotes the word-spotting without noise.**

## 4.8. Experiment with Date Spotting in Other Dataset

### 4.8.1. ICDAR dataset

The proposed method was evaluated on the ICDAR 2013 Robust Reading competition [23] dataset. This dataset contains 913 images. We have collected word images from this dataset and next we have generated a total of 2252 line images two to six word images in a line. We considered 214 line images as validation, 425 as testing and rest as training. A list of 100 words was considered for word spotting using our framework of word spotting. Binary image and gray image features are combined using a Tandem framework. Next shape coding based approach was integrated in Tandem-HMM combination. Fig. 20 shows the precision-recall curve of word spotting performance without and with shape coding approach. Shape coding integration in word spotting outperforms the plain Tandem combination.

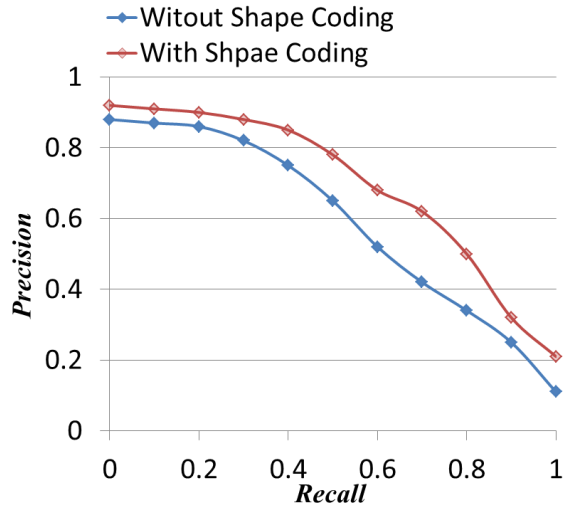

**Fig. 20 Word spotting performance on ICDAR dataset using out shape coding based approach along with tandem feature.**

### 4.8.2. Non-Latin Script

To test the robustness of our proposed framework, we have considered handwritten documents of Indic (Devanagari and Bangla) scripts for performance evaluation of shape coding based date spotting approach. We have collected text-line images of both Devanagari and Bangla script. Some of these line images contain date field in English script. We considered 514 and 498 line images as testing for Devanagari and Bangla respectively. No script-wise training was performed for Latin date spotting in these text lines. Some examples of qualitative results are shown in Fig.21. We obtained encouraging results using our framework. For quantitative measures, an average of the results from 3 different date models is taken. We have obtained 68.12 and 69.47 as global MAP values for Bangla and Devanagari script respectively.

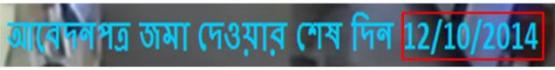

Fig. 21: Date spotting performance on Bangla and Devanagari script.

## 4.8. Time computation

Experiments have been done on an I5 CPU of 2.80 GHz and 4G RAM 664 Computer. For each query, the average runtime has been computed from different runs made in the experiment. In Table VI, we show the time taken using queries of various approaches developed in Matlab. By using shape coding the time for date spotting is reduced to 1.16 seconds which is close to similar performance using binary image features.

Table VI: Time computation analysis of date spotting using different approaches.

| Date Spotting Approach | Running Time |
|---|---|
| Binary Image Feature | 1.12 Sec |
| Gray Image Feature | 1.07 Sec |
| Binary + Gray Image Feature | 1.32 Sec |
| Binary + Gray Image Feature + Tandem | 1.49 Sec |
| Binary + Gray Image Feature + Tandem + Shape Coding | 1. 16 Sec |

## 5. Conclusion

In this paper, we proposed an efficient system for spotting dates in video frames without recognizing it. First we presented a novel feature extraction method for text characters using image enhancement by wavelet and gradient features. Combination of binary image feature and gray image feature using Tandem approach improves the HMM-based word spotting performance significantly. To further boost the performance, a shape coding based approach is used to combine the similar shape characters in same class during word spotting. It reduces the character confusion and hence improves the spotting performance. The date-spotting framework has been extended to locate Latin date information in other scripts. Proposed framework has been tested in a large dataset of scene images and video frames. We obtained encouraging results using this framework.